\documentclass[letterpaper, 10 pt, conference]{ieeeconf} 
\IEEEoverridecommandlockouts
\usepackage{graphicx}
\usepackage{multirow}
\usepackage{pbox}
\usepackage{xcolor}
\usepackage[super]{nth}
\usepackage{amsmath}
\usepackage{tikz}
\usepackage{textcomp}
\usepackage{url}
\usepackage{hyperref}
\usepackage[hyphenbreaks]{breakurl}

\title{\LARGE \bf Scalable Supervisory Architecture for Autonomous Race Cars}

\author{Zalán Demeter, Péter Bogdán, Ármin Bogár-Németh, Gergely Bári%
\thanks{Széchenyi István University, H-9026 Győr, Hungary (e-mail: zalan.demeter@humda.hu)}%
}

\newcommand\copyrighttext{%
  \footnotesize \textcopyright 2024 IEEE. Personal use of this material is permitted. Permission from IEEE must be obtained for all other uses, in any current or future media, including reprinting/republishing this material for advertising or promotional purposes, creating new collective works, for resale or redistribution to servers or lists, or reuse of any copyrighted component of this work in other works.
  DOI: \href{https://doi.org/10.1109/IV55156.2024.10588615}{10.1109/IV55156.2024.10588615}}
\newcommand\copyrightnotice{%
\begin{tikzpicture}[remember picture,overlay]
\node[anchor=south,yshift=10pt] at (current page.south) {\fbox{\parbox{\dimexpr\textwidth-\fboxsep-\fboxrule\relax}{\copyrighttext}}};
\end{tikzpicture}%
}

\begin{document}

\maketitle
\copyrightnotice
\thispagestyle{empty}
\pagestyle{empty}

\begin{abstract}

In recent years, the number and importance of autonomous racing leagues, and consequently the number of studies on them, has been growing. The seamless integration between different series has gained attention due to the scene's diversity. However, the high cost of full scale racing makes it a more accessible development model, to research at smaller form factors and scale up the achieved results. This paper presents a scalable architecture designed for autonomous racing that emphasizes modularity, adaptability to diverse configurations, and the ability to supervise parallel execution of pipelines that allows the use of different dynamic strategies. The system showcased consistent racing performance across different environments, demonstrated through successful participation in two relevant competitions. The results confirm the architecture's scalability and versatility, providing a robust foundation for the development of competitive autonomous racing systems. The successful application in real-world scenarios validates its practical effectiveness and highlights its potential for future advancements in autonomous racing technology.

\end{abstract}

\section{Introduction}
\label{sec:introduction}

The concept of self-driving vehicles has been present for a considerable duration. Since the beginning of the 1900s, there has been a fascination with the potential of autonomous vehicles. In the present day, this fascination and enthusiasm remain strong. Numerous companies are investing significant amounts of money into researching and developing such systems. However, the true intricacy of the subject is continually being revealed, as the projected timeline for the availability of highly automated vehicles is consistently being delayed.

Motorsport has always been at the forefront of developing new technologies for the automotive industry, and self-driving racing is no exception too \cite{Betz_2022}. There have already been many competitions in the past, like the DARPA Grand Challenge \cite{Darpa2004} and there are ongoing challenges and race series also today, like the 1/\nth{10} scale F1Tenth \cite{OKelly2019}, Roborace \cite{Roborace2020}, the Indy Autonomous Challenge (IAC) \cite{IAC2021} and the recently announced Abu Dhabi Racing League (A2RL) \cite{A2RL2024}.

The automotive industry is going through a fundamental change. From the distributed embedded systems, it is slowly moving towards centralized architectures. Software is becoming the main differentiating and value adding part of the whole system. Compared to the old hardware-driven approach, this requires a different skill and mindset of the involved parties \cite{Ernts2023}. 

This is why software is going to be the main component in any future autonomous vehicle, and good software starts with a good architecture. In this paper we propose a novel scalable system architecture which can be used in multiple different racing vehicles (e.g. simulation based, 1/\nth{10} scale, 1/\nth{3} scale, full-scale formula race car, etc.). Moreover, this architecture is capable of running multiple agents (drivers) simultaneously compared to most of the current architectures, which allow only one agent (driver) to control the vehicle.
Starting development in simulation environment and later working up from smaller scale cars to real size vehicles is a cost efficient way to start developing autonomous vehicles. Having a scalable software architecture that can be continuously used and evolved during this period can reduce the overall effort and resources needed to design and build such systems.

\section{Overview of Autonomous Race car architectures}
\label{sec:overview_architectures}

\begin{figure}[!ht]
 \centering
 \includegraphics[width=8.0cm]{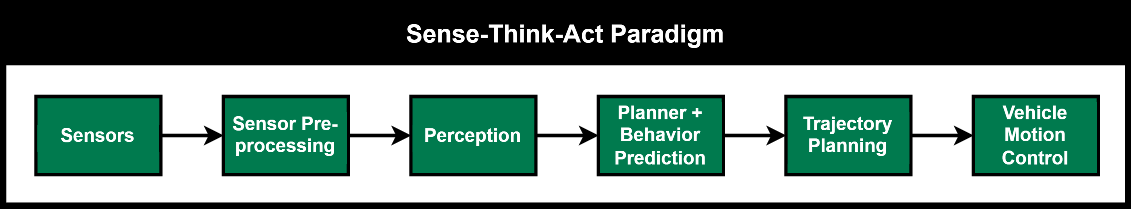}
 \caption{Sense-Think-Act Paradigm: Designing software architecture using this paradigm involves environment perception, processing information, and taking actions based on analysis.}
 \label{fig:sensethinkact}
\end{figure}

In both the automotive industry and the motorsports context, a widely used software architecture for autonomous vehicles follows the Sense-Think-Act paradigm. A simplified architecture overview using this paradigm can be seen in \hyperref[fig:sensethinkact]{Figure \ref{fig:sensethinkact}}.

A similar architecture has been used in the Indy Autonomous Challenge by both the TUM Autonomous Motorsport team \cite{Betz2022} and the TII EuroRacing team \cite{Raji2023}.

Usually the sensors include LiDAR sensors, video cameras, radar sensors, ultrasonic sensors, satellite-based positioning sensors, inertial measurement units, wheel speed sensors and additional vehicle-specific sensors.

These sensors continuously provide the necessary information about both the environment and the ego vehicle. Information is collected about the vehicle's surroundings, other vehicles, road infrastructure, and obstacles which are then processed and analyzed by the perception module where they are filtered, tracked, classified, and localized. The output of the perception subsystem is a comprehensive environmental model.

These environmental data are processed and interpreted by the system where sophisticated algorithms make predictions, plans and decisions based on either predefined rules, machine learning models, or other inputs like signals coming from race control. These decisions and plans will become the basis of a trajectory planner which creates and selects the best possible future trajectory that the vehicles intend to follow. This trajectory is based not only on one single objective (e.g. maximum speed), but also has to consider physical limitations and safety considerations as well.

Finally, a vehicle motion module will execute the received trajectory and control the vehicle via the steering, accelerator, and brake actuators, while ensure an effective and safe operation. Then the whole cycle starts again.

In addition to the functional part of the architecture, there are also modules which serve the non-functional requirements. These include failure detection, resource management, degradation concept, system state monitoring, data recording, and live telemetry. These modules usually connect to many or, in some cases, to all other modules.

\begin{figure}[!ht]
 \centering
 \includegraphics[width=8.0cm]{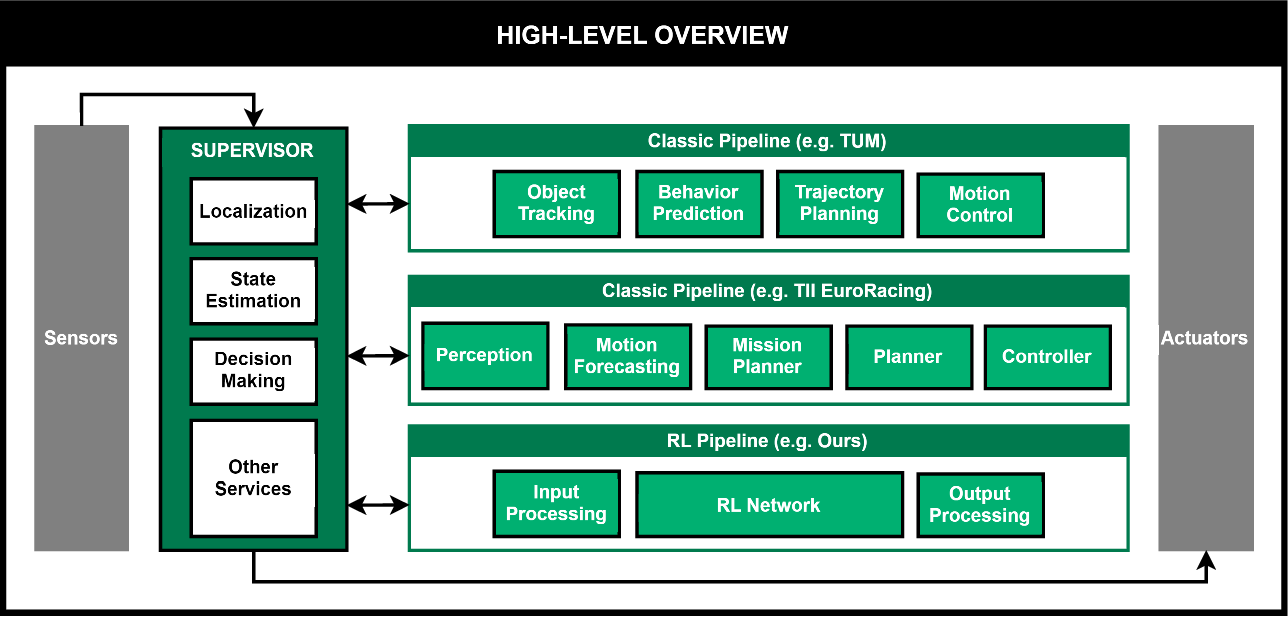}
 \caption{High Level Overview: example of multiple pipelines working together in the architecture.}
 \label{fig:highleveloverview}
\end{figure}

In the architecture of both IAC teams, a common trait is that the architecture is intertwined with the driving agent. This means that there is only one agent running at any given time. In this paper, we propose an architecture that allows us to run multiple driving agents (also referred to as "pipelines"). This is necessary to allow multiple agents to run parallel, as in recent years there have been significant advancements in artificial intelligence research of autonomous vehicles. These advancements encompass a wide range of AI based technologies. Starting from small AI algorithms, such as data-driven approaches in the perception, through partial end-to-end learning based methods and finally full end-to-end AI driven systems. These new approaches must coexist with the currently existing classical approaches because there are certain situations where one approach works better than the other. This also helps to have an efficient development, testing, deployment, and reusability of multiple approaches. Moreover, such an architecture reduces the effort needed to reuse it on different platforms. A high-level example of multiple pipelines used in the architecture can be seen in \hyperref[fig:highleveloverview]{Figure \ref{fig:highleveloverview}}. A detailed description of the architecture will be discussed in \hyperref[sec:overview]{Section \ref{sec:overview}}.

\section{Architecture Overview}
\label{sec:overview}

\begin{figure*}[!ht]
 \centering
 \includegraphics[width=16.0cm]{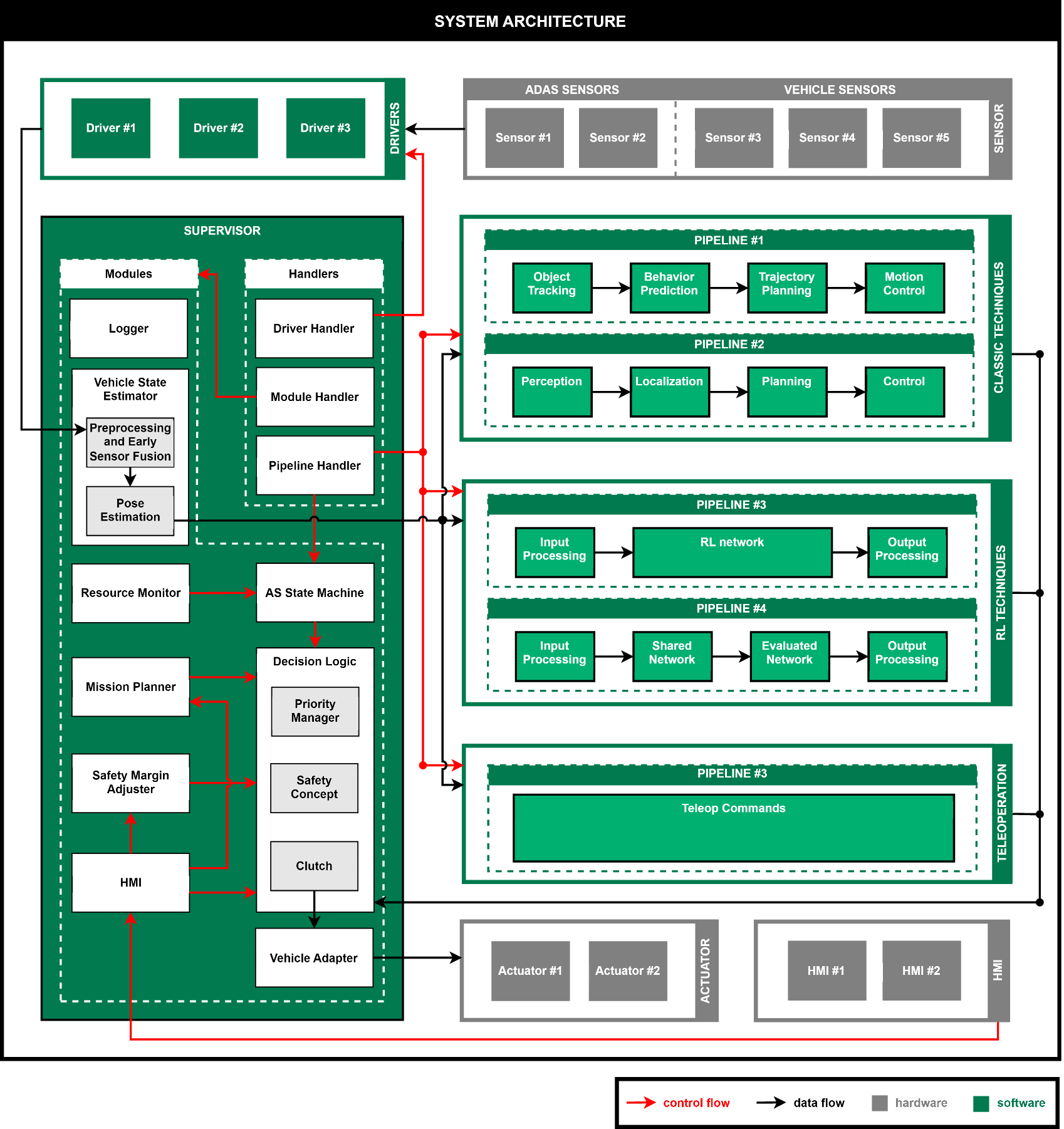}
 \caption{Architecture Diagram: Data and control flow of the system. The components depicted are detailed in \hyperref[sec:modules]{Section \ref{sec:modules}}, \hyperref[sec:handlers]{Section \ref{sec:handlers}} and \hyperref[sec:pipelines]{Section \ref{sec:pipelines}}.}
 \label{fig:arch}
\end{figure*}

\subsection{Driving Factors}
\label{sec:drivingfactors}
The architectural design is based on several drivers that affect the core functionality and the implemented components of the system. During the design process, four main factors were identified, namely: reusability, scalability, ability to run parallel pipelines and the supervisory nature of the system. As described in \hyperref[sec:introduction]{Section \ref{sec:introduction}} due to the changing nature of autonomous racing leagues the ability to reuse previously developed algorithms in different environments and scenarios is vital. Different racing leagues use different racing formats that require partially different algorithms, thus well defined interfaces are needed to allow for interchangeability of specific components and adapters to various vehicles. Due to resource and funding constraints, algorithm testing usually starts on smaller, cheaper platforms such as the aforementioned F1/10 \cite{OKelly2019} format. This emphasises the need for the architecture to support migration and scalability from smaller systems to 1/1 scale complex autonomous vehicles. As the scale grows from 1/\nth{10} - F1/10 \cite{OKelly2019} to 1/\nth{3} - Go-Kart \cite{Qiao2023} to full scale - A2RL \cite{A2RL2024} the autonomous hardware and sensor stack inherently becomes extensive and compound which calls for the need to implement multiple state-of-the-art autonomous techniques. Taking this into account, algorithms of any complexity should be easy to integrate into the system, and they should be able to work in parallel and redundantly. Hence the need for a supervisory system, as the algorithm must be able to handle the results of multiple policies and make reasoned decisions between them.

\subsection{Components of the architecture}
\label{sec:components}
\begin{figure}[!ht]
 \centering
 \includegraphics[width=8.0cm]{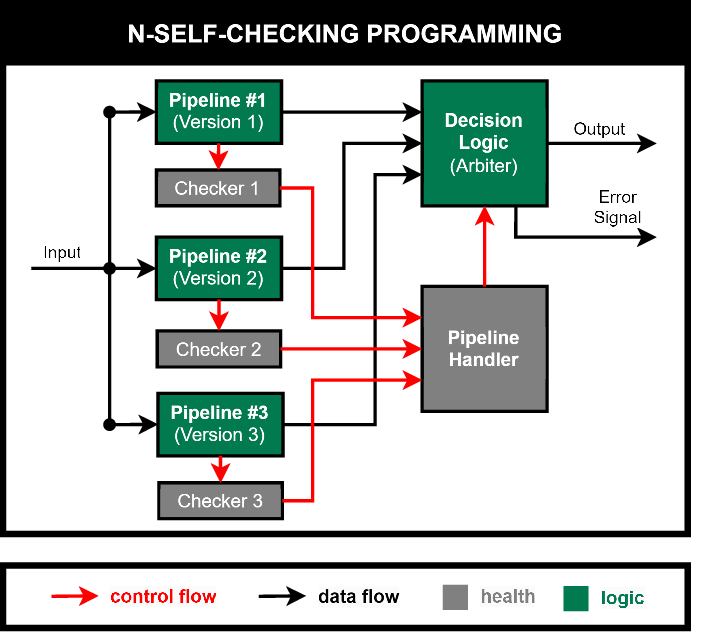}
 \caption{N-self-checking programming: concurrently operating multiple independent versions with checkers for fault tolerance. This is an example configuration, the depicted components are detailed in \hyperref[sec:decisionlogic]{Section \ref{sec:decisionlogic}} and \hyperref[sec:pipelinehandler]{Section \ref{sec:pipelinehandler}}.}
 \label{fig:nselfchecking}
\end{figure}

The architecture consists of different hardware and software modules, each serving distinct functional elements. The data flow relationship between them will be presented in \hyperref[sec:dataflow]{Section \ref{sec:dataflow}} and the governing/control process will be discussed in \hyperref[sec:controlflow]{Section \ref{sec:controlflow}}. The hardware components will not be presented in this article, the system developed is sensor and actuator independent and can be applied to any vehicle by developing the appropriate interface. In later articles, the full hardware system will be described, as the distributed nature of the components is also of great importance. The first software component is the driver, which is a critical interface unit. However, for the reasons previously stated, we will not go into detail about its individual functionalities. The two main logical units are the set of pipelines and the supervisory system itself. The relationship of the former with general architectures has already been highlighted in \hyperref[sec:overview_architectures]{Section \ref{sec:overview_architectures}}. Each pipeline executes an independent computational process to calculate the best action for the current state with potential prior information included. This falls in line with the concept of N-self-checking programming which has the important role of introducing fault tolerance with the use of active dynamic redundancy. This design structure is depicted in \hyperref[fig:nselfchecking]{Figure \ref{fig:nselfchecking}}. The supervisor is the core of our architecture, a miniature operating system that monitors and controls the subsystems and all running processes in our autonomous environment. Its main functionalities are serving as a monitoring system (resource acquisition control, data logging, software health monitoring), launching and life-cycle handling required drivers, modules, and pipelines and acting as the main decider and output transition handler. These features are composed of modules and handlers, which logically separate the control of internal and external functional units. The modules realize the services provided by the supervisor, while the handlers mainly act as runners and managers for external software units, also including this functionality for the internal modules. This allows for a unified interface for all software components governed by the supervisor. All modules and handlers referenced here are later described in \hyperref[sec:modules]{Section \ref{sec:modules}} and \hyperref[sec:handlers]{Section \ref{sec:handlers}}.

\subsection{Data Flow}
\label{sec:dataflow}
\begin{figure}[!ht]
 \centering
 \includegraphics[width=8.0cm]{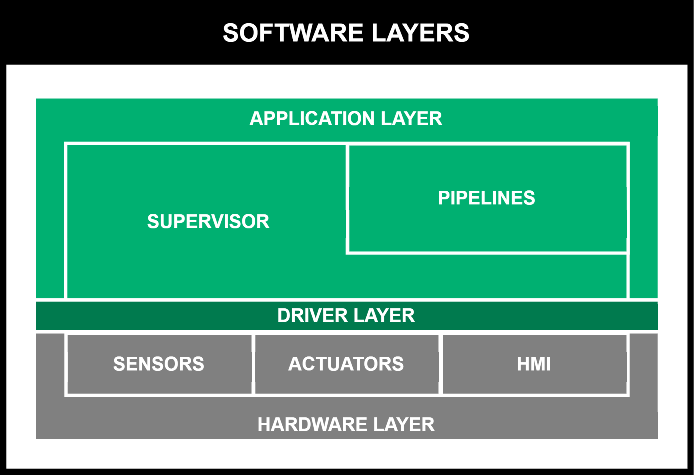}
 \caption{Software Layer Diagram: shows the hierarchy between functional layers in the system.}
 \label{fig:swlayers}
\end{figure}

\begin{figure}[!ht]
 \centering
 \includegraphics[width=8.0cm]{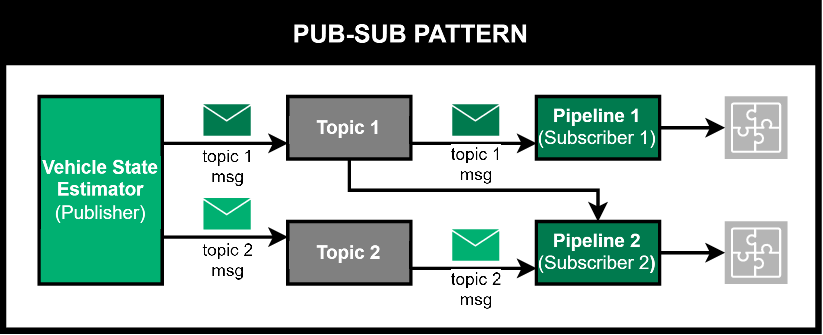}
 \caption{Pub Sub Pattern: asynchronous messaging in flexible and scalable software. This is an example configuration, the depicted components are detailed in \hyperref[sec:vehiclestatestimator]{Section \ref{sec:vehiclestatestimator}} and \hyperref[sec:pipelines]{Section \ref{sec:pipelines}}.}
  \label{fig:pubsub}
\end{figure}

The transfer of data between various components or modules of the system is called data flow. It displays how data is processed and changed within the system by showing the route that it travels from its source to its destination. The complete flow of data from its origin through processing and up to actuation can be followed in \hyperref[fig:arch]{Figure \ref{fig:arch}}.  By the layered nature of the software system, the data travel through different layers of execution. These layers are depicted in \hyperref[fig:swlayers]{Figure \ref{fig:swlayers}}. First, sensors produce raw or partially processed data in a variety of formats and quantities in the sensor layer, depending on vehicles and race formats. These data are given and cannot be interchanged with data coming from different sensor configurations. Then the driver layer processes the incoming stream of data into a comprehensible and well-defined data formats that can be directly used by the application layer. The different components of the Application Layer referenced here are detailed in \hyperref[sec:modules]{Section \ref{sec:modules}} and \hyperref[sec:handlers]{Section \ref{sec:handlers}}. The data packets received by the application layer reach the pipelines through a unified processing unit, the Vehicle State Estimator. Every pipeline has access to the required data streams specified by itself. This is made possible with the use of the Pub-Sub pattern shown in \hyperref[fig:pubsub]{Figure \ref{fig:pubsub}}, which is generally achieved by, but not exclusive to ROS \cite{Macenski2022} . The pipelines collect the input data and through various processes and calculations, produce control signals and corresponding probability density functions (PDFs) describing the correctness of the output. The PDFs play a crucial role as part of the control flow and will be discussed in \hyperref[sec:controlflow]{Section \ref{sec:controlflow}}. The control output vector is the direct input to the decision-making process carried out by the Decision Logic module. After the decision has been made, the application to the appropriate vehicle is carried out by the Vehicle Adapter module, which plays an important role in the reusability of the developed architecture. While the input to this module is the last unified interface of the data flow, the output interface connects the software to the actuators, which serve as the last sink for the data traveling throughout the system.

\subsection{Control Flow}
\label{sec:controlflow}
The sequential execution of instructions within a program, influencing its behavior and order of operations, is known as control flow. It shows the paths the program follows when it comes to conditions and factors that affect its behavior. The different control paths and the interconnection of the components can be followed in \hyperref[fig:arch]{Figure \ref{fig:arch}}. The first functional block when it comes to controlling the environment is the set of handlers that manage the instantiation and life-cycle handling of different internal and external software components. They decide what to run and when, so they form the core of the decision-making control system. The Pipeline Handler plays a prominent role among the handlers, as it receives information about the state of each pipeline instance as a PDF mentioned in (Section 2). These PDFs, combined with information from having full ownership of the pipelines, make for a good representation of the health of the car's driving agents. The other coherent unit of control is the Autonomous System State Machine (AS State Machine) and the Decision Logic module, which have multiple different input modules. The AS State Machine collects information on the state of the system from all relevant units of implementation and identifies possible states and transitions between them. This module tells the Decision Logic module what is feasible at a certain time. In addition to the aforementioned, the decision logic collects information from the Human Machine Interface (HMI) and connected to this the Safety Margin Adjuster and the Mission Planner. These additional inputs all serve different purposes that will be discussed later. The Decision Logic is the last and most crucial element of the Supervisor's control flow. It receives information from all areas of the autonomous system and makes a rational, robust decision based on what it determines to be the most appropriate and safest at the time. External factors influencing the decision making process of the architecture can be supplemented by an arbitrary number of inputs using the well-specified interface of the HMI. 


\section{Handlers}
\label{sec:handlers}
\begin{figure}[!ht]
 \centering
 \includegraphics[width=8.0cm]{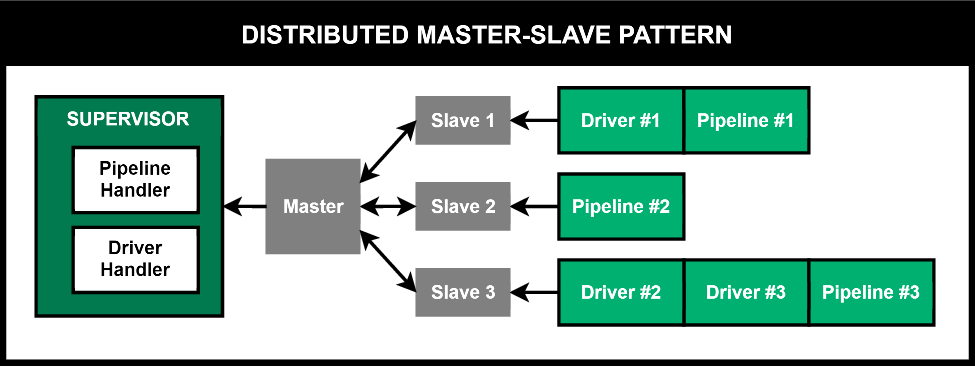}
 \caption{Distributed Master-Slave Pattern: for parallel processing and aggregate results, a central master node assigns and coordinates tasks to several independent slave nodes. This is an example configuration, the depicted components are detailed in \hyperref[sec:pipelinehandler]{Section \ref{sec:pipelinehandler}} and \hyperref[sec:driverhandler]{Section \ref{sec:driverhandler}}.}
 \label{fig:masterslave}
\end{figure}

Handlers are central to the process of instantiating and running a specific configuration of the software stack. They provide the system with a high degree of flexibility, which can be used to operate the different internal and external software units in a uniform environment on a wide range of systems with varying setup options. State-of-the-art autonomous techniques have high computational requirements, and running them parallel is even more demanding. Different sensors may also require separate processing units due to possible I/O and bandwidth limitations. This poses the need for system distribution to multiple different processing units. The proposed structure allows the architecture to use the distributed master-slave pattern also supported by, but not exclusive to ROS \cite{Macenski2022}. An example distributed configuration of the architecture is shown in \hyperref[fig:masterslave]{Figure \ref{fig:masterslave}}.

\subsection{Driver Handler}
\label{sec:driverhandler}
The Driver Handler acts as the main originator and master for the set of drivers required for a specific vehicle. The components managed by it can be easily interchanged for any type of vehicle configuration. Its responsibility is to launch every driver for the sensors and actuators required. It has information about the current hardware configuration and the distributed nature of the compute units. In addition, it can stop selected drivers on request, due to resource limitations, or start requested drivers if the current race situation requires so.

\subsection{Module Handler}
\label{sec:modulehandler}
The Module Handler is an abstract concept, responsible for managing and launching internal modules. The reason for its existence is to have the same interface as the handlers of external software components and to be able to reconfigure the internal operation of the supervisor, depending on the services needed in a given environment. This allows supervisory processes such as resource monitoring, lifecycle management,and watchdog functions to be implemented in a unified way.

\subsection{Pipeline Handler}
\label{sec:pipelinehandler}
The Pipeline Handler plays an important role in the group of handlers. In addition to launching, managing, and monitoring the pipelines currently needed in the system, possibly in a distributed configuration, it also has the responsibility to maintain the health state of the pipeline instances it administers and to communicate it to the state monitoring system and indirectly to the decision logic. By using the principle of appropriate software design patterns, no information is required for the software stack of a given pipeline. It can handle different implementations on a common interface. When a new pipeline is inserted, only one pipeline class needs to be implemented, and the new solution fits into the existing system.

\section{Modules}
\label{sec:modules}
Modules are a collection of services implemented by the supervisor. They are versatile units that cover many different areas. Their common trait is that they play a prominent role in the supervision of the autonomous system and in the establishment of common interfaces.

\subsection{Vehicle State Estimator}
\label{sec:vehiclestatestimator}
One of the main aspects of architecture design was to be able to run pipelines with multiple approaches in parallel. The Vehicle State Estimator is a key element of this design concept. A wide range of state-of-the-art autonomous algorithms require a diverse set of inputs. But it is possible to identify algorithmic units that may be required by more than one method, so there are advantages to externalizing them into a common subsystem. One of the benefits is that there is no need to create and execute the same algorithm twice, which comes with runtime benefits and a lower chance of software errors. This does not mean that the presented structure inevitably creates Single Point of Failures (SPOFs). Due to the fact that multiple autonomous algorithms can be implemented in the system, a deep knowledge and analysis of the system can be used to design pipelines that do not depend solely on common inputs. In addition, if necessary, software redundancy can be built into the module itself and into the processing algorithms running in the module, all a matter of design choices. As can be seen in \hyperref[fig:arch]{Figure \ref{fig:arch}} we propose to place the position estimation (localization), preprocessing of signals and early sensor fusion algorithms here. It serves as a single source of data for the pipelines, because each driver agent accesses the set of data it requires through it. This module is responsible for implementing a core concept, namely the easy integration of different vehicles. It is easily interchangeable and can be customized to the sensor configuration requirements of the specified vehicle.

\subsection{Vehicle Adapter}
\label{sec:vehicleadapter}
The Vehicle Adapter module is the direct counterpart of the Vehicle State Estimator introduced in \hyperref[sec:vehiclestatestimator]{Section \ref{sec:vehiclestatestimator}}. While the latter unifies the interface with the hardware on the input side, this module performs the same task on the output side. The input to this module is the control signal, i.e. the representation of the action to be performed at a given moment in time, in normalized form. This signal consists of two components, the  \( steering \ angle \in [-1, 1]\) and the  \(pedal \ position \in [-1, 1]\). These values are normalized representations of the minimum and maximum steering angle and the maximum throttle and maximum brake command, under the condition that the driver agent does not brake and accelerate at the same time. The architecture we propose abstracts the vehicle dynamics control (VDC) functional unit and the low level control system from the autonomous logic, this means that it can handle both electric and internal combustion powertrains in a uniform way. Therefore, it takes as a prerequisite the functions supported by the vehicles, such as automatic transmission (AT), anti-lock braking system (ABS), electronic stability control (ESC), etc.

\subsection{AS State Machine}
\label{sec:asstatemachine}
The task of the autonomous system state machine is to know and collect all the states of the system, to define possible state transitions and their associated conditions. For the proper functioning of a well-structured system, it is essential to confine software operations to a given operational framework. Here we can define the fundamental safety unit between the self-driving vehicle and those around it. We can describe when the car can enter autonomous mode, and define safety conditions in case of software failure. And it is also possible to indicate and manage run-time problems.  The actual implementation is not discussed, as it is not specific to the architecture and can be chosen arbitrarily to fit the safety, integrity  and reliability requirements.

\subsection{Decision Logic}
\label{sec:decisionlogic}
Decision Logic is one of the most important and central units of our proposed architecture. All the factors affecting the movement of the vehicle and its environment get processed here. The sole instrument for decision making between actions planned by all redundant subsystems. As mentioned in \hyperref[sec:controlflow]{Section \ref{sec:controlflow}}, as part of the control flow, the PDFs for the control signal and the health state associated with the pipelines are also evaluated here. As shown in \hyperref[fig:arch]{Figure \ref{fig:arch}}, this complex unit is broken down into three submodules. The priority manager's task is to determine the static or dynamic priority among the evaluated pipelines in response to the current race situation or external intervention. Static priority can be set via the HMI or parameterised for specific race modes. The logic of dynamic priority can be chosen in any way desired by designing appropriate objective functions. The correct choice of this is not negligible, as it critically affects system security. Therefore, the status of another subsystem should be taken into account, namely the Safety Concept. Its function is to determine when the output of a pipeline is safe to use in the current system state. To give an example, the robustness of Reinforcement Learning (RL) agents is not guaranteed by themselves, but their use can bring significant benefits. Therefore, in addition to using an RL pipeline, it may be necessary to use a classical solution method, such as a robust control. It is the responsibility of the Safety Concept to decide whether a safe state would result from the RL agent's intended intervening signal or whether it is necessary to switch back to the classic pipeline as part of a safety maneuver. This can be accompanied, if necessary, by a Degradation Concept submodule, which is responsible for managing the safety risks caused by physical degradation associated with the vehicle (e.g. tyres). In addition to the above, another critical submodule is the Clutch, which is responsible for managing the transition between pipelines. The transition between control methods is by no means a given and is an area of active research. When switching between control methods, it is necessary to consider whether switching from a state that is considered robust by the current controller allows one to reach a robust state associated with another controller. The actual implementation of this is a complex subject in itself and will be covered in a future paper.

\subsection{Resource Monitor}
\label{sec:resourcemonitor}
The purpose of the Resource Monitor module is to provide a service to query the runtime status and requirements of all units active in the system. It is also responsible for estimating resources and overseeing life cycle management. It works in close cooperation with AS State Machine and is an integral part of the security subsystem.The real implementation is highly environment and operating system dependent, so it may be necessary to create multiple instances depending on the requirements.

\subsection{Mission Planner}
\label{sec:missionplanner}
It is the Mission Planner's responsibility to collate the information needed to implement the race events. It is specific to the race format and can be easily changed according to current needs. It can also include general maneuvers such as in- or out laps, etc. It is also possible to abstract general manoeuvres from specific details for a common implementation.

\subsection{Human Machine Interface}
\label{sec:humanmachineinterface}
The Human Machine Interface consists of a software unit that receives and processes signals from the hardware communication platform. The required intervention functions can be communicated to the system through a uniform and defined interface. The hardware implementation also varies, and there are several different solutions. For this reason, this unit can also be easily exchanged and adapted to the circumstances. It allows us to fit a given hardware equipment to the functions we want to implement.

\subsection{Safety Margin Adjuster}
\label{sec:safetymarginadjuster}
We define self-driving algorithms as driver agents. In interpretation, we bring software drivers as close as possible to real racing drivers, this has several advantages. This allows experiences and processes from our motorsport background to be understood and applied to self-driving agents, the Safety Margin Adjuster module is a prime example of this. Race engineers are responsible for managing the risk-taking of racing drivers, and this module gives them the opportunity to do so. From the pit wall it is possible to influence the safety margin of the system via the HMI. The logic of choosing the appropriate risk-taking strategy in response to an external command can be defined in this subsystem.

\subsection{Logger}
\label{sec:logger}
Saving and analyzing data is an essential tool to develop a self-driving system and evaluate its competitive performance. For this reason, the Logger module was included, whose special task is to save the necessary software versions and parameter sets for every event that took place on the track in a traceable and recoverable way. Its operation is automatic and can be connected to the AS State Machine. If the system enters a specified state, the Logger automatically starts its operation and logs the data until the vehicle reaches the end state. The Logger can be supplemented with any required function and its implementation can also be chosen arbitrarily.

\section{Pipelines}
\label{sec:pipelines}
Pipelines are the set of algorithms that control the autonomous vehicle. Each pipeline can be considered as a driver agent capable of performing the entire set of autonomous functions on its own. Implementations can range widely and use a wide variety of input sensor data. We distinguish two fundamentally different solutions, the classical group of methods and the learning-based solutions. The task of classical pipelines is to produce reliable and predictable intervention signals with a proven method besides the not always guaranteed correct signals of learning-based algorithms. In addition to these two categories, we distinguish a third pipeline category with a special role. The Teleoperation Pipelines are responsible for the complete remote control of the vehicle. In addition, certain predefined control tasks can be implemented here, such as reversing maneuvers, lane change manoeuvres or even a skidpad test mode or any arbitrary control task. It is a guiding principle of the architecture that these methods can be implemented arbitrarily and easily integrated into the architecture. Therefore it can serve as a convenient platform for racing and a standard foundation for research on self-driving algorithms. The Classic and RL pipelines we have developed will be discussed in detail in separate articles \cite{classic2024} \cite{rl2024}. 

\section{RESULTS}
\label{sec:results}
In order to demonstrate the applicability and scalability of the proposed architecture in different environments, we present two competition events where we have run the system successfully and analyse the results. We participated in the competitions presented in \hyperref[sec:F1TenthIROS23]{Section \ref{sec:F1TenthIROS23}} and \hyperref[sec:IACSimRace2023]{Section \ref{sec:IACSimRace2023}}.

\subsection{F1Tenth IROS23 Competition}
\label{sec:F1TenthIROS23}
\begin{figure}[!ht]
 \centering
 \includegraphics[width=8.0cm]{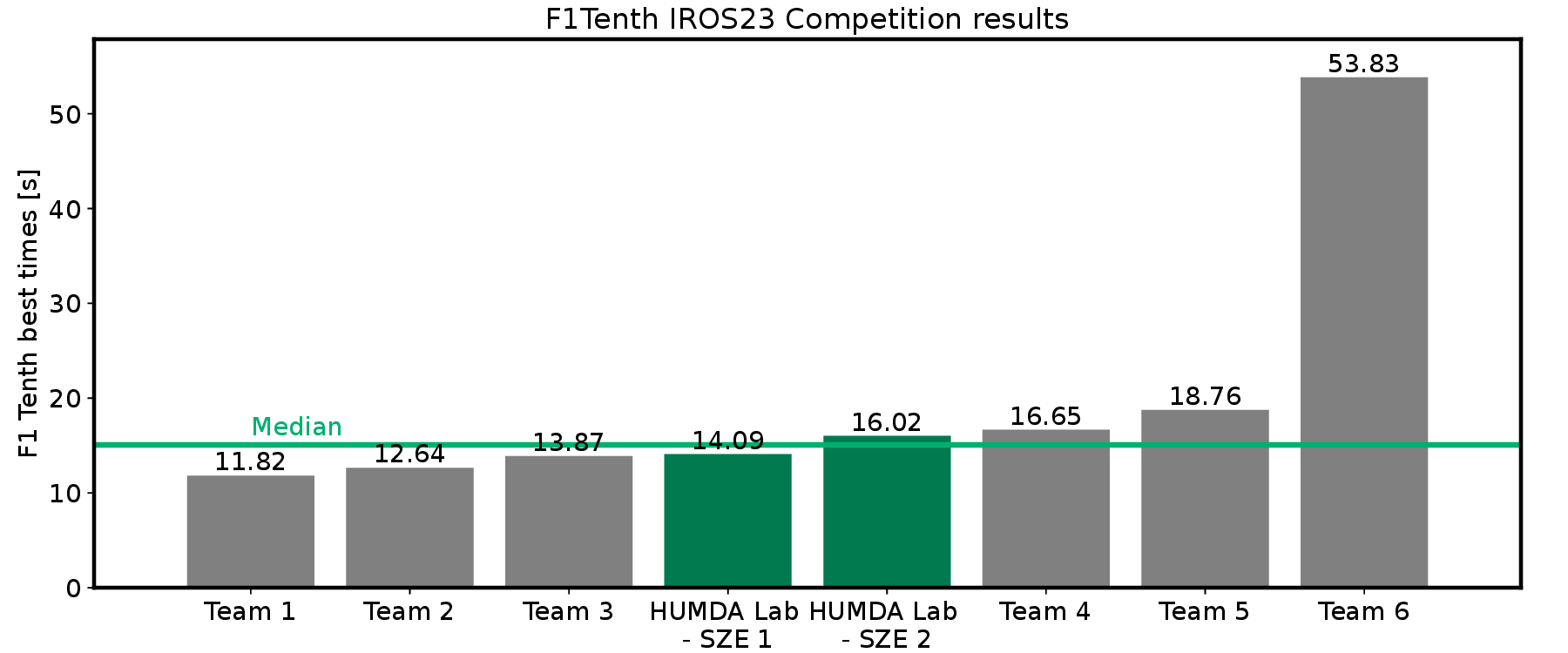}
 \caption{F1Tenth IROS23 Competition results: the best lap times of the finishing teams and their median. The teams we entered and thus using the architecture are highlighted in green.}
  \label{fig:f1tenth}
\end{figure}

IROS23 F1tenth \cite{OKelly2019} competition which was held in person at the IROS23 conference in Detroit, MI, United States from October \nth{1} to \nth{5}, 2023. The competition was held over three days. The first day was free practice, second day was qualification day and race was held on the last day. In total nine teams from all over the world participated in the event. Eight of these teams managed to achieve a measured lap time, two of which were set by our team. The results can be reviewed in \hyperref[fig:f1tenth]{Figure \ref{fig:f1tenth}}. 
\newline
\par
For the first time, the architecture we designed was translated into reality at this event. The platform cars used by the competition have a simple sensor set, so the integration of the driver layer consisted of installing the HMI (PlayStation 4 controller) driver, a LiDAR sensor driver and the motor controller driver. The Vehicle State Estimator module has been adapted to this sensor system and the corresponding signals and processing have been implemented. We ran an instance of each category of pipeline presented in \hyperref[sec:pipelines]{Section \ref{sec:pipelines}}. A classic pipeline based on local planning and geometric control, a deep reinforcement learning based Proximal Policy Optimization (PPO) pipeline, and a teleop pipeline for full remote control of the car. The output of these normalised and unified interface compliant signals was converted to the necessary control signal set for vehicle actuation using a Vehicle Adapter to match the vehicle hardware configuration.

\subsection{Indy Autonomous Challenge Simulation Race 2023}
\label{sec:IACSimRace2023}
\begin{figure}[!ht]
 \centering
 \includegraphics[width=8.0cm]{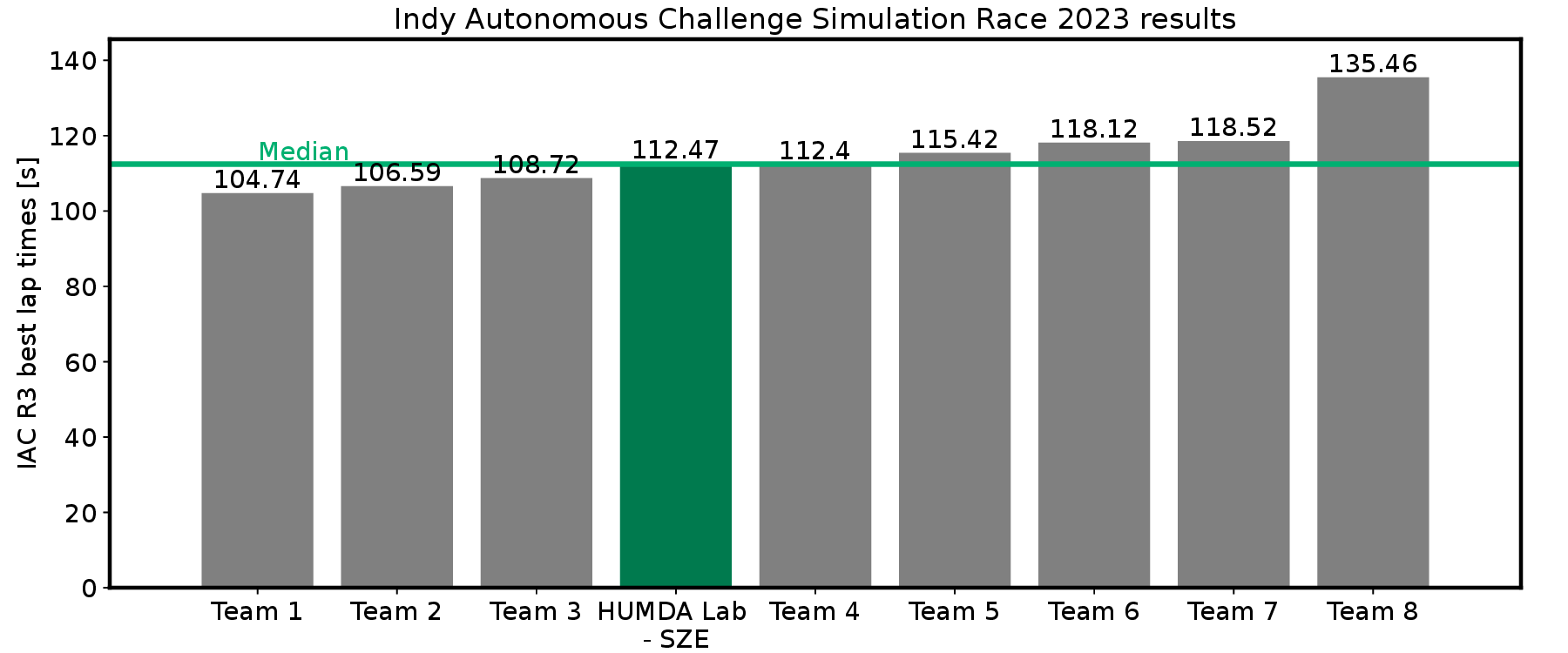}
 \caption{Indy Autonomous Challenge Simulation Race 2023 results: the best lap times of the finishing teams and their median. The teams we entered and thus using the architecture are highlighted in green.}
  \label{fig:iac}
\end{figure}

Indy Autonomous Challenge \cite{IAC2021} Simulation Race was held in the fall of 2023 as a virtual simulation event. The event used the simulator provided by Autonoma \cite{Autonoma2022}. It took place virtually on the famous Autodromo Nazionale Monza also known as the "Temple of speed". The vehicle used was a model based on the vehicle used in the real life Indy Autonomous : the Dallara AV-21R \cite{DallaraAV21}. It consisted of three rounds separated by two weeks, where the teams had the opportunity to develop and tune their algorithms. The three rounds had an added level of complexity. The first round had a "time attack" format, where the competitors had to complete one warm-up lap followed by two flying laps on an empty track to set the fastest total time. The second challenge added additional non-playable vehicles, which the contestant had to overtake to achieve the same number of laps as round one. The last round had the same format as round two, with the added difficulty of simulated bad measurements and dropouts of the GPS signals of the ego vehicle modeled after real-world experience from the real IAC race in MIMO 2023 \cite{Mimo2023}. The end result was the final results of round three. In total, 11 teams submitted to at least one of the rounds, but only nine teams submitted to the final round. The results can be reviewed in \hyperref[fig:iac]{Figure \ref{fig:iac}}. 
\newline
\par
During our participation in the F1Tenth race, we developed a software stack, divided into pipelines, which we were able to apply to this race as well, making the appropriate additions and modifications. The car model in the simulation environment had completely different characteristics, sensor settings, and actuation. However, with the help of the architecture, we were able to integrate our existing implementation into a completely new system in a very short time. Due to the nature of the simulation, the driver layer is not included in this configuration. Regardless, we had to implement the appropriate Vehicle Adapter to provide the input for the pipelines, and the output actuation interface for the Vehicle Adapter had to be adapted to the environment.

\section{CONCLUSIONS}
We have integrated the proposed architecture into two different environments and have been able to participate successfully in the respective competition series as a result. To demonstrate usability and scalability, we will examine and quantify the results achieved. We used essentially the same principle of self-driving pipelines in both races, so the driver agent's racing performance is generic between the two race events. A significant difficulty was the difference between the two environments, which the proposed architecture helped to bridge. By examining the lap times, we can see that in both races our team's result is at the median of the lap times (\hyperref[fig:f1tenth]{Figure \ref{fig:f1tenth}} and \hyperref[fig:iac]{Figure \ref{fig:iac}}). This demonstrates that the race performance of our driver agent was indeed consistent between events. This confirms that the architecture is scalable, as we have successfully scaled up our entire system from a 1/\nth{10} scale vehicle to a simulated but nevertheless full-scale race car. In addition, evidence of invariance to environments (real and simulated) and the ease of reusability of pipelines is also provided by the results presented above. In the F1Tenth competition, we also showed the ability to run multiple parallel pipelines and dynamically prioritize between them using a simple objective function. These have been used to demonstrate, through real-life tests, that the requirements set out in the architectural design have been met. In the coming period, our team will use this architecture in the A2RL \cite{A2RL2024} racing series, so we will also test its operation on a real full-scale racing car. In order to further expand on the topic, we will later present the entire internal behavior of the supervisory system, including the Autonomous System State Machine, the Decision Logic, and its important submodules, including the Clutch, among others.

\addtolength{\textheight}{-12cm}

\section*{ACKNOWLEDGMENT}
The research was supported by the National Research, Development and Innovation Office. (2020-2.1.1-ED-2021-00162)


\bibliographystyle{IEEEtran}
\bibliography{references}

\begin{thebibliography}{10}
\providecommand{\url}[1]{#1}
\csname url@samestyle\endcsname
\providecommand{\newblock}{\relax}
\providecommand{\bibinfo}[2]{#2}
\providecommand{\BIBentrySTDinterwordspacing}{\spaceskip=0pt\relax}
\providecommand{\BIBentryALTinterwordstretchfactor}{4}
\providecommand{\BIBentryALTinterwordspacing}{\spaceskip=\fontdimen2\font plus
\BIBentryALTinterwordstretchfactor\fontdimen3\font minus
  \fontdimen4\font\relax}
\providecommand{\BIBforeignlanguage}[2]{{%
\expandafter\ifx\csname l@#1\endcsname\relax
\typeout{** WARNING: IEEEtran.bst: No hyphenation pattern has been}%
\typeout{** loaded for the language `#1'. Using the pattern for}%
\typeout{** the default language instead.}%
\else
\language=\csname l@#1\endcsname
\fi
#2}}
\providecommand{\BIBdecl}{\relax}
\BIBdecl

\bibitem{Betz_2022}
\BIBentryALTinterwordspacing
J.~Betz, H.~Zheng, A.~Liniger, U.~Rosolia, P.~Karle, M.~Behl, V.~Krovi, and
  R.~Mangharam, ``Autonomous vehicles on the edge: A survey on autonomous
  vehicle racing,'' \emph{IEEE Open Journal of Intelligent Transportation
  Systems}, vol.~3, p. 458–488, 2022. [Online]. Available:
  \url{http://dx.doi.org/10.1109/ojits.2022.3181510}
\BIBentrySTDinterwordspacing

\bibitem{Darpa2004}
\BIBentryALTinterwordspacing
{\relax Defense Advanced Research Projects Agency}. (2005) Darpa the grand
  challenge. [Online]. Available:
  \url{https://www.darpa.mil/about-us/timeline/-grand-challenge-for-autonomous-vehicles}
\BIBentrySTDinterwordspacing

\bibitem{OKelly2019}
M.~O'Kelly, V.~Sukhil, H.~Abbas, J.~Harkins, C.~Kao, Y.~V. Pant, R.~Mangharam,
  D.~Agarwal, M.~Behl, P.~Burgio, and M.~Bertogna, ``F1/10: An open-source
  autonomous cyber-physical platform,'' 2019.

\bibitem{Roborace2020}
\BIBentryALTinterwordspacing
{\relax Roborace}. (2020) Roborace. [Online]. Available:
  \url{https://roborace.com/}
\BIBentrySTDinterwordspacing

\bibitem{IAC2021}
\BIBentryALTinterwordspacing
{\relax Indy Autonomous Challenge}. (2021) Indy autonomous challenge. [Online].
  Available: \url{https://www.indyautonomouschallenge.com/}
\BIBentrySTDinterwordspacing

\bibitem{A2RL2024}
\BIBentryALTinterwordspacing
{\relax ASPIRE}. (2024) Abu dhabi autonomous racing league in uae. [Online].
  Available: \url{https://a2rl.io/}
\BIBentrySTDinterwordspacing

\bibitem{Ernts2023}
\BIBentryALTinterwordspacing
{\relax Ernst \& Young}. (2023) The software-driven revolution redefining the
  automotive industry. [Online]. Available:
  \url{https://assets.ey.com/content/dam/ey-sites/ey-com/en_in/topics/automotive-and-transportation/2023/ey-the-software-driven-revolution-redefining-the-automotive-industry.pdf}
\BIBentrySTDinterwordspacing

\bibitem{Betz2022}
J.~Betz, T.~Betz, F.~Fent, M.~Geisslinger, A.~Heilmeier, L.~Hermansdorfer,
  T.~Herrmann, S.~Huch, P.~Karle, M.~Lienkamp, B.~Lohmann, F.~Nobis,
  L.~Ögretmen, M.~Rowold, F.~Sauerbeck, T.~Stahl, R.~Trauth, F.~Werner, and
  A.~Wischnewski, ``Tum autonomous motorsport: An autonomous racing software
  for the indy autonomous challenge,'' \emph{Journal of Field Robotics}, 2022.

\bibitem{Raji2023}
A.~Raji, D.~Caporale, F.~Gatti, A.~Giove, M.~Verucchi, D.~Malatesta, N.~Musiu,
  A.~Toschi, S.~R. Popitanu, F.~Bagni, M.~Bosi, A.~Liniger, M.~Bertogna,
  D.~Morra, F.~Amerotti, L.~Bartoli, F.~Martello, and R.~Porta, ``er.autopilot
  1.0: The full autonomous stack for oval racing at high speeds,'' 2023.

\bibitem{Qiao2023}
Z.~Qiao, M.~Zhou, T.~Agarwal, Z.~Zhuang, F.~Jahncke, P.-J. Wang, J.~Friedman,
  H.~Lai, D.~Sahu, T.~Nagy, M.~Endler, J.~Schlessman, and R.~Mangharam,
  ``Av4ev: Open-source modular autonomous electric vehicle platform to make
  mobility research accessible,'' 2023.

\bibitem{Macenski2022}
S.~Macenski, T.~Foote, B.~Gerkey, C.~Lalancette, and W.~Woodall, ``Robot
  operating system 2: Design, architecture, and uses in the wild,''
  \emph{Science Robotics}, vol.~7, no.~66, May 2022.

\bibitem{classic2024}
M.~Fazekas, Z.~Demeter, J.~Tóth, Ármin Bogár-Németh, and G.~Bári,
  ``Evaluation of local planner-based stanley control in autonomous rc car
  racing series,'' in \emph{2024 IEEE Intelligent Vehicles Symposium (IV)},
  2024, pp. 252--257.

\bibitem{rl2024}
M.~Hell, G.~Hajgató, Ármin Bogár-Németh, and G.~Bári, ``A lidar-based
  approach to autonomous racing with model-free reinforcement learning,'' in
  \emph{2024 IEEE Intelligent Vehicles Symposium (IV)}, 2024, pp. 258--263.

\bibitem{Autonoma2022}
\BIBentryALTinterwordspacing
{\relax Autonoma}. (2022) Scalable simulation tools for safer autonomy.
  [Online]. Available: \url{https://www.autonomalabs.com/}
\BIBentrySTDinterwordspacing

\bibitem{DallaraAV21}
\BIBentryALTinterwordspacing
{\relax Dallara}. (2021) The av-21 racecar. [Online]. Available:
  \url{https://www.indyautonomouschallenge.com/racecar}
\BIBentrySTDinterwordspacing

\bibitem{Mimo2023}
\BIBentryALTinterwordspacing
{\relax Milano Monza Motor Show}. (2023) Indy autonomous challenge, the
  unmanned cars premiere at mimo 2023. [Online]. Available:
  \url{https://www.milanomonza.com/en/event-info}
\BIBentrySTDinterwordspacing

\end{thebibliography}

\end{document}